\pdfoutput=1

\documentclass[11pt]{article}

\usepackage[final]{acl}

\usepackage{times}
\usepackage{latexsym}

\usepackage[T1]{fontenc}

\usepackage[utf8]{inputenc}

\usepackage{microtype}

\usepackage{inconsolata}

\usepackage{graphicx}

\usepackage{amsmath,amssymb,amsfonts}
\usepackage{amsthm}
\usepackage{booktabs}
\usepackage{multicol}
\usepackage{multirow}
\usepackage{color}
\usepackage{xcolor}
\usepackage{subfigure}

\usepackage{algorithm}
\usepackage{algorithmic}

\title{

PPC-GPT: Federated Task-Specific Compression of Large Language Models via Pruning and Chain-of-Thought Distillation
}

\author{
 \textbf{Tao Fan\textsuperscript{1, 2}},
 \textbf{Guoqiang Ma\textsuperscript{2}},
 \textbf{Yuanfeng Song\textsuperscript{2}},
 \textbf{Lixin Fan\textsuperscript{2}},
 \textbf{Qiang Yang\textsuperscript{3}}
\\
 \textsuperscript{1}Hong Kong University of Science and Technology, Hong Kong, China
 \\
 \textsuperscript{2}WeBank Co., Ltd, Shenzhen, China
 \\
 \textsuperscript{3}Hong Kong Polytechnic University, Hong Kong, China
\\
 \small{
   \textbf{Correspondence:} \href{mailto:tfanac@cse.ust.hk}{tfanac@cse.ust.hk}, \href{mailto:qyang@cse.ust.hk}{qyang@cse.ust.hk}
 }
}

\begin{document}
\maketitle

\begin{abstract}

Compressing Large Language Models (LLMs) into task-specific Small Language Models (SLMs) encounters two significant challenges: safeguarding domain-specific knowledge privacy and managing limited resources. To tackle these challenges, we propose PPC-GPT, a novel unified framework that systematically addresses both privacy preservation and model compression in federated settings. PPC-GPT works on a server-client federated architecture, where the client sends differentially private (DP) perturbed task-specific data to the server's LLM. The LLM then generates synthetic data along with their corresponding rationales. This synthetic data is subsequently used for both LLM pruning and retraining processes. Our framework's key innovation lies in its holistic integration of privacy-preserving mechanisms, synthetic data generation, and task-specific compression techniques, creating unique benefits through component interaction. Our experiments across diverse text generation tasks demonstrate that PPC-GPT successfully achieves dual objectives: maintaining competitive performance comparable to full-sized LLMs while ensuring robust privacy protection through its federated architecture. Our code has been contributed to the FATE open-source project and is now publicly accessible at \textit{\url{https://github.com/FederatedAI/FATE-LLM/tree/main/python/fate_llm/algo/ppc-gpt}}

\end{abstract}

\section{Introduction}

Large Language Models (LLMs), such as GPT-4~\cite{Gpt-4} and LLaMA3-70B~\cite{dubey2024llama}, boasting billions of parameters and remarkable text generation capabilities, have emerged as a transformative force in the realm of artificial intelligence. However, their training demands substantial computational resources~\cite{openai2023gpt4tr}, and their colossal size poses significant hurdles for practical deployment, especially in resource-limited environments.
Conversely, Small Language Models (SLMs), such as OPT-1.3B~\cite{zhang2022opt} and Qwen2.5-1.5B~\cite{team2024qwen2}, frequently demonstrate superior computational efficiency and accelerated response rates, making them ideally suited for real-time applications with constrained resources. Enterprises with constrained resources typically prefer deploying SLMs, as they can do so without the concern of potential data leaks, a risk that is heightened when utilizing remote LLMs~\cite{fan2025ten,fan2025fedmkt,fan2023fate-llm,kang2023grounding}.
Yet, training an SLM from scratch, even the smallest billion-parameter models, entails considerable computational expenses that are financially prohibitive for most enterprises. Furthermore, SLMs exhibit inherent limitations that stem from their performance constraints.

In this work, we aim to tackle the following question: \textit{Is it feasible to develop a task-specific and competitive SLM by harnessing an existing pre-trained LLM for enterprises with limited resources, while ensuring compliance with privacy requirements?}
To achieve this objective, we delve into structured pruning~\cite{xia2024sheared, men2024shortgpt, kim2024shortened}, as a viable approach. Pruning is generally regarded as a strategy for compressing task-specific models by eliminating redundant parameters and expediting inference, all while maintaining task performance.

We identify two crucial technical challenges associated with this problem:
Firstly, how can we ensure the privacy of task-specific data when enterprises with limited resources are unable to prune an LLM into an SLM independently? In such cases, the need to transmit task-specific data to a remote server equipped with powerful computing resources arises, a practice that is frequently unacceptable to most enterprises due to privacy concerns.
Secondly, how can we ensure that the performance of the SLM remains comparable to that of the LLM? Structured pruning inevitably leads to some degree of performance degradation.
To overcome these challenges, we introduce PPC-GPT, a privacy-preserving federated framework designed for compressing LLMs into task-specific SLMs via pruning and Chain-of-Thought (CoT) distillation.

\begin{figure*}[ht]
 \centering

  \includegraphics[width=0.8\linewidth]{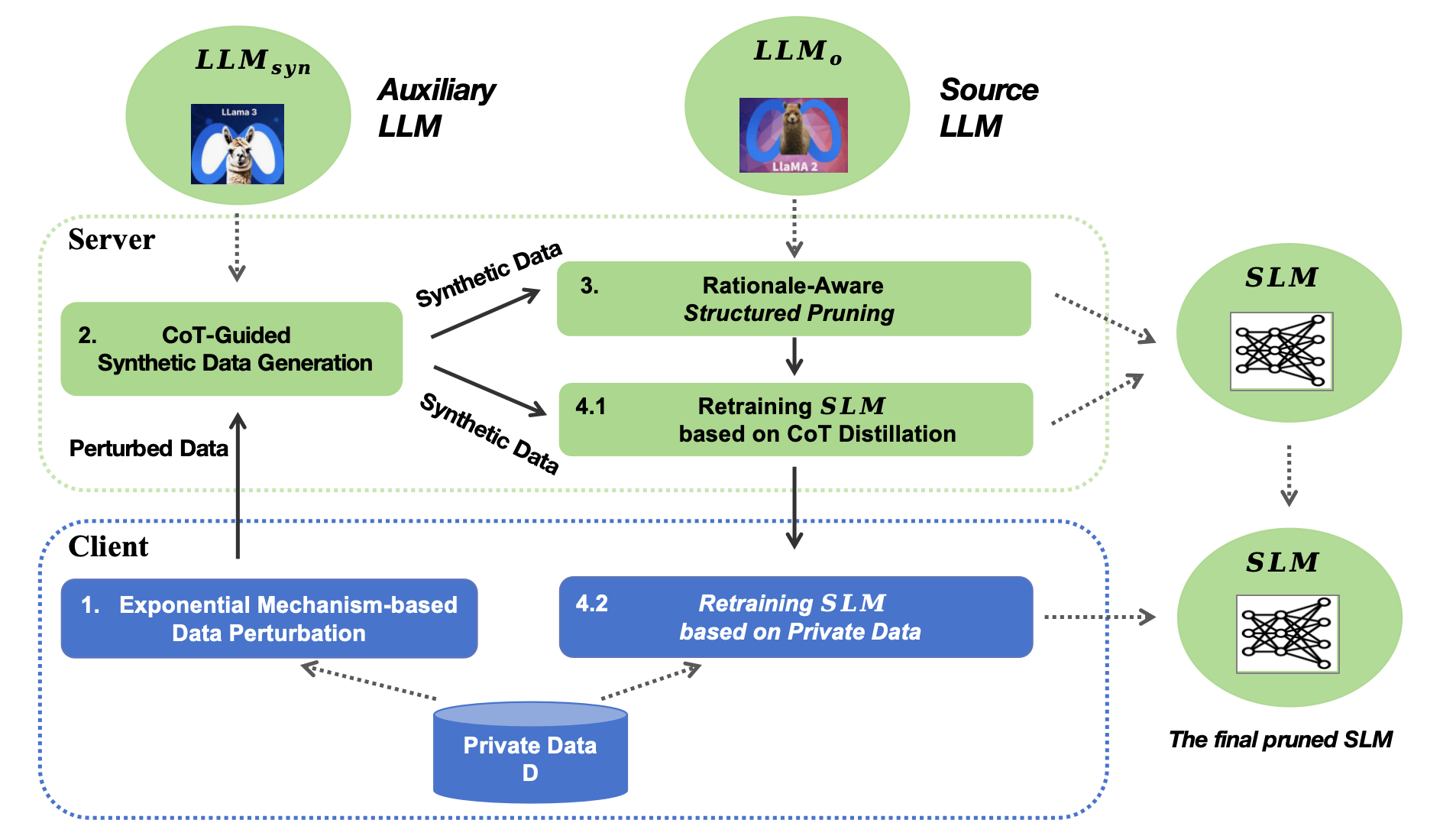}
  
    \caption{The overview of our proposed \textbf{PPC-GPT}. The PPC-GPT comprises four key components: (1) The \textit{Exponential Mechanism-based Data Perturbation}, which perturbs the client's data to ensure privacy;  (2) The \textit{CoT-Guided Synthetic Data Generation}, responsible for creating new synthetic data and rationales based on the perturbed data; (3) The \textit{Rationale-Aware Structured Pruning}, a process that prunes original $LLM_o$ to obtain the target smaller $SLM$;  (4) The \textit{Retraining SLM}, where the target $SLM$ is retrained using both synthetic and original private data to restore accuracy.}
 \label{fig:ppc-gpt}

\end{figure*}

As depicted in Figure \ref{fig:ppc-gpt}, the envisioned architecture of PPC-GPT comprises a high-performance server adept at deploying LLMs and facilitating their pruning into SLMs, coupled with a client endowed with more constrained computational capabilities for running SLMs.
Within the confines of our framework, the workflow unfolds as detailed below. Initially, the client sends task-specific data, perturbed to ensure privacy, to the server. These data are protected by the Exponential Mechanism of Differential Privacy~\cite{dwork2006differential, mcsherry2007mechanism,tong2025inferdpt}, thereby guaranteeing privacy protection. Subsequently, the server-side auxiliary $LLM_{syn}$ generates synthetic data along with their corresponding rationales, based on these perturbed inputs.
The server-side $LLM_o$, which represents the original model, undergoes pruning by PPC-GPT to yield the \textit{target SLM}. This pruning process is informed by both the synthetic data and their associated rationales. Following the pruning of the $LLM_o$, the server retrains the \textit{target SLM} through CoT ~\cite{wei2022chain, hsieh2023distilling, li2023symbolic} knowledge distillation, leveraging the same synthetic data and rationales. Lastly, the server dispatches the refined \textit{target SLM} to the client, who then proceeds to retrain the \textit{target SLM} utilizing its locally private data.

Our contributions can be summarized as follows:

\begin{itemize}

\item 
\textbf{Unified Framework for Federated LLM Compression.} We propose PPC-GPT, the first holistic framework designed to systematically unify privacy-preserving techniques with LLM compression in a federated learning context. Its core innovation lies in the synergistic integration of four key modules: exponential mechanism-based data perturbation, CoT-guided synthetic data generation, rationale-aware structured pruning, and CoT-based knowledge distillation. This modular architecture is not only novel but also extensible, allowing for future advancements in any of its constituent parts.

\item 
\textbf{Novel Rationale-Aware Structured Pruning for Reasoning Preservation.} 
We propose a new structured pruning metric that evaluates network layers based on their contribution to generating explanatory rationales. This approach selectively preserves the model's essential reasoning capabilities, a critical aspect often overlooked in standard compression techniques.

\item 
\textbf{Component Interaction Benefits.} The integration creates unique advantages through component interaction. For example, combining DP-perturbed data with CoT-guided synthetic data generation enables both privacy protection and effective knowledge transfer, while the rationale-aware structured pruning leverages this enhanced data quality for better compression decisions.

\item 
\textbf{Empirical Assessment of LLM Compressing to Task-Specific SLM.} Through extensive experiments across various text generation tasks using LLaMA and OPT models, we demonstrate how PPC-GPT's component interactions lead to effective task-specific compression while maintaining privacy, achieving results competitive with full-sized LLMs.

\end{itemize}

\section{Related Work}

\subsection{Differential Privacy}
\label{sec:dp}
In this section, We briefly revisit two important definitions of differential privacy: $\epsilon$-Differential Privacy and Exponential Mechanism (EM).

\textbf{$\epsilon$-Differential Privacy (DP)}.
\textit{The Definition of $\epsilon$-Differential Privacy (DP)}~\cite{dwork2006differential}. 
A randomized algorithm $M: D \to S $ is $\epsilon$-Differential Privacy if for any two neighboring datasets $D_1, D_2 \in D$ that differ exactly in a single data sample, and for any output $O \subseteq S$:

\begin{equation}\label{eq:dp-0}
\begin{aligned}
     P_r[M(D_1) \in O] \le e^{\epsilon} P_r[M(D_2) \in O]
\end{aligned}
\end{equation}
where $\epsilon$ is a privacy parameter. Smaller values of $\epsilon$ imply stronger privacy guarantees.

\textbf{Exponential Mechanism}.
\textit{The Definition of Exponential Mechanism}~\cite{mcsherry2007mechanism, tong2025inferdpt}. For a given scoring function $u: X\times Y \to  R$, a randomized mechanism $M(X, u, Y)$ is $\epsilon$-DP compliant if it satisfies: 

\begin{equation}\label{eq:dp-1}
\begin{aligned}
     P_r[y|x]\propto exp(\frac{\epsilon \cdot u(x,y) }{2\bigtriangleup u }  )
\end{aligned}
\end{equation}
where the sensitivity $\bigtriangleup u$ is defined as:
\begin{equation}\label{eq:dp-2}
\begin{aligned}
    \bigtriangleup u = \max_{x,x^{'} \in X, y\in Y} |u(x,y) - u(x^{'},y)| 
\end{aligned}
\end{equation}

\subsection{Differential Privacy Synthetic Data}

A practical approach to generating private synthetic data involves training a language model, such as LLaMA2-7B~\cite{touvron2023llama2}, on private data using DP through DP-SGD~\cite{song2013stochastic,bassily2014private,abadi2016deep}. Subsequently, the DP model is sampled repeatedly to produce synthetic data ~\cite{mattern2022differentially,yue-etal-2023-synthetic,kurakin2023harnessing}. 
Research conducted by~\cite{mattern2022differentially,yue-etal-2023-synthetic,kurakin2023harnessing} demonstrates that training downstream models on DP synthetic data achieves performance comparable to training directly on real data with DP, thereby underscoring the high quality of the synthetic data.

However, a significant challenge arises because cutting-edge LLMs, like GPT-4, do not offer model weights, making DP fine-tuning impractical. Even for open-source LLMs, such as LLaMA3-70B~\cite{dubey2024llama}, the process is resource-intensive. 
Meanwhile, these DP fine-tuning methods inherently rely on a trusted server to gather data from data owners for model training~\cite{chen-etal-2023-customized}, significantly limiting their applicability in scenarios where such trusted servers are not available, as is the case in our research context.
In the context of this work, we operate within a client-server architecture where fine-tuning the LLM on the server is not an option.

\subsection{Model Pruning}

Model pruning, initially proposed by~\cite{lecun1989optimal} and subsequently enhanced by~\cite{han2015learning}, stands as a resilient and efficient strategy for mitigating model redundancy and attaining compression. This methodology branches into two primary techniques: \textit{unstructured pruning and structured pruning}.

Unstructured pruning ~\cite{dong2017learning,lee2018snip,wang2020picking,sun2024a,frantar2023sparsegpt} can obtain highly compressed models by directly pruning neurons, disregarding the model's internal architecture, which also causes unstructured sparsity and hard deployment.
A more pragmatic and structured option is structured pruning. \textit{Structured pruning} targets organized patterns for removal, encompassing entire layers~\cite{jha2023train}, attention heads within Multi-Head Attention (MHA) mechanisms~\cite{michel2019sixteen}, hidden sizes in Feedforward Neural Networks (FFN)~\cite{nova2023gradient}, as well as hybrid configurations~\cite{kurtic2024ziplm}. 
In recent times, there has been a surge in structured pruning research tailored specifically for LLMs. For example, ShortGPT~\cite{men2024shortgpt}, LaCo~\cite{yang2024laco}, and Shortened LLaMA~\cite{kim2024shortened} concentrate solely on pruning depth (i.e., layer-wise). LLM-Pruner~\cite{ma2023llm} eliminates coupled structures in relation to network width while preserving the layer count. Sheared-LLaMA~\cite{xia2024sheared} introduces a mask learning phase that is designed to pinpoint prunable components in both network width and depth.
Our work falls in the category of structured pruning of LLMs.

\section{Problem Formulation}
\label{sec:problem}

Given an LLM $f_{\theta} $ with parameters $ \theta$, which represents the original LLM that requires pruning, and a task-specific dataset $ \mathcal{D} $ containing private data, our objective is to develop a target smaller, task-specific compressed SLM $f_{\phi} $ parameterized by $ \phi $. To acheive this,  we seek to find the optimal pruning strategy $ \mathcal{P} $ and retraining approach $ \mathcal{R} $. The objective can be formulated as follows:
\begin{equation}
\begin{aligned}
& \quad \quad \quad \quad \min_{\mathcal{P}, \mathcal{R}} \mathcal{L}(\phi; \theta,\mathcal{D}) \\
& s.t. \quad \left| \phi \right| \ll \left| \theta \right| \ \quad \text{and} \quad \mathcal{L}_{p}(\mathcal{D})<\delta
\end{aligned}
\label{eq:formulation}
\end{equation}
where $ \mathcal{L}(\phi; \theta,\mathcal{D}) $ is the loss function measuring the performance of the compressed SLM on the task-specific dataset. $ |\phi| $ and $ |\theta| $ denote the number of parameters in the compressed and original models, respectively. $ \mathcal{L}_{p}(\mathcal{D}) $ is the privacy loss incurred due to the perturbation of the data to ensure differential privacy.

Our goal is to find the optimal pruning strategy $ \mathcal{P} $ and retraining approach $ \mathcal{R} $ that minimizes the overall loss, taking into account both the performance of the compressed SLM and the privacy protection of the task-specific data in the client. We assume the server to be \textit{semi-honest}, meaning it may attempt to extract the client's private data from the information it receives.

\section{The Proposed PPC-GPT Framework}

In this section, we introduce PPC-GPT, a unified privacy-preserving federated framework for compressing LLMs into task-specific SLMs. 
We illustrate the PPC-GPT architecture in Figure~\ref{fig:ppc-gpt}. As detailed in Algorithm~\ref{alg:ppc-gpt}, our framework comprises four key modules that work in concert:

\begin{itemize}
    \item \textit{Exponential Mechanism-based Data Perturbation}: Ensures privacy protection through exponential mechanism.
    \item \textit{CoT-guided Synthetic Data Generation}: Creates high-quality synthetic data and rationales.
    \item \textit{Rationale-Aware Structured Pruning}: Leverages synthetic data and rationales for model compression.
    \item \textit{Retraining SLM}: Optimizes the compressed model through two-stage knowledge distillation.
\end{itemize}

We elaborate on these modules in Section \ref{sec:perturb}, \ref{sec:synthetic}, \ref{sec:pruning} and \ref{sec:retraining}, respectively. Through this integrated approach, PPC-GPT effectively addresses the challenges of privacy-preserving LLM compression while maintaining task-specific performance.

\begin{algorithm}[t]
\caption{PPC-GPT}
\label{alg:ppc-gpt}
\begin{algorithmic}[1]
\REQUIRE Private dataset $\mathcal{D} = \{(x_i, y_i)\}_{i=1}^N$, $LLM_{syn}$ for synthetic data generation, Original $LLM_{o}$ $f_{\theta}$ that requires pruning, Privacy budget $\epsilon$
\ENSURE Task-specific SLM $f_{\phi}$ with privacy guarantees

\STATE \textbf{Phase 1: Exponential Mechanism-based Data Perturbation}
\STATE Apply Exponential Mechanism $\mathcal{M}$ to $\mathcal{D}$ with budget $\epsilon$
\STATE Generate $\mathcal{D}_p = \{(x^p_i)\}_{i=1}^N$ according to Eq.(\ref{eq:perturb})

\STATE \textbf{Phase 2: CoT-guided Synthetic Data Generation}
\STATE Server's $LLM_{syn}$ processes $\mathcal{D}_p$ to generate synthetic data
\STATE Generate $\mathcal{D}_s = \{(x^s_i, (y^s_i, r^s_i))\}_{i=1}^{N_s}$

\STATE \textbf{Phase 3: Rationale-Aware Structured Pruning}
\STATE Calculate Block Influence scores using $\mathcal{D}_s$ according to Eq.(\ref{eq:bi})
\STATE Identify redundant layers based on BI values
\STATE Obtain pruned model structure $f_{\phi}$ where $|\phi| \ll |\theta|$

\STATE \textbf{Phase 4: Retraining SLM via Two-Stage Knowledge Distillation}
\STATE Server performs CoT distillation using $\mathcal{D}_s$ according to Eq.(\ref{eq:cot_kd}), (\ref{eq:ft}), (\ref{eq:rationale})
\STATE Client fine-tunes with private data $\mathcal{D}$ according to Eq.(\ref{eq:ft_client})

\RETURN Compressed task-specific SLM $f_{\phi}$
\end{algorithmic}
\end{algorithm}

\subsection{Exponential Mechanism-based Data Perturbation}
\label{sec:perturb}
We utilize an exponential mechanism~\cite{mcsherry2007mechanism,yue2021differential,chen-etal-2023-customized} to perturb the local private data $\mathcal{D} = \left\{ (x_i, y_i) \right\}_{i=1}^N$, which satisfies the criteria for the $\epsilon$-DP. 
For detailed information about the exponential mechanism, please refer to Section \ref{sec:dp}. We denote the  perturbed dataset as $\mathcal{D}_p= \left\{ (x^p_i) \right\}_{i=1}^N$, where $x^p_i$ signifies an perturbed input based on the original local private dataset $\mathcal{D}$ .

The Exponential Mechanism $ \mathcal{M} $ is defined as a randomized algorithm that, given the original local private dataset $ \mathcal{D} $, outputs the perturbed dataset $ \mathcal{D}_p $ with probability proportional to the exponential of the utility score (in this work, we use cosine similarity as the utility
function):
\begin{equation}
\begin{aligned}
\mathcal{M}(\mathcal{D}) = \mathcal{D}_p \quad \text{with prob} \propto exp(\frac{\epsilon \cdot u(\mathcal{D},\mathcal{D}_p) }{2\bigtriangleup u }  )
\end{aligned}
\label{eq:perturb}
\end{equation}

\subsection{CoT-guided Synthetic Data Generation}
\label{sec:synthetic}
When the server-side $LLM_{syn}$ receives the perturbed data $\mathcal{D}_p$, the server initiates a procedure where $LLM_{syn}$ generates fresh synthetic data along with their corresponding rationales based on these perturbed data. 
We denote the synthetic dataset as $\mathcal{D}_s = \left\{ (x^s_i, (y^s_i, r^s_i)) \right\}_{i=1}^{N_s}$, where $x^s_i$ signifies an input, $y^s_i$ signifies the corresponding expected output label, $r^s_i$ signifies the desired rationale, and $N_s$ represents the sample size of synthetic data. 

\begin{figure}[ht]
 \centering

  \includegraphics[width=1\linewidth]{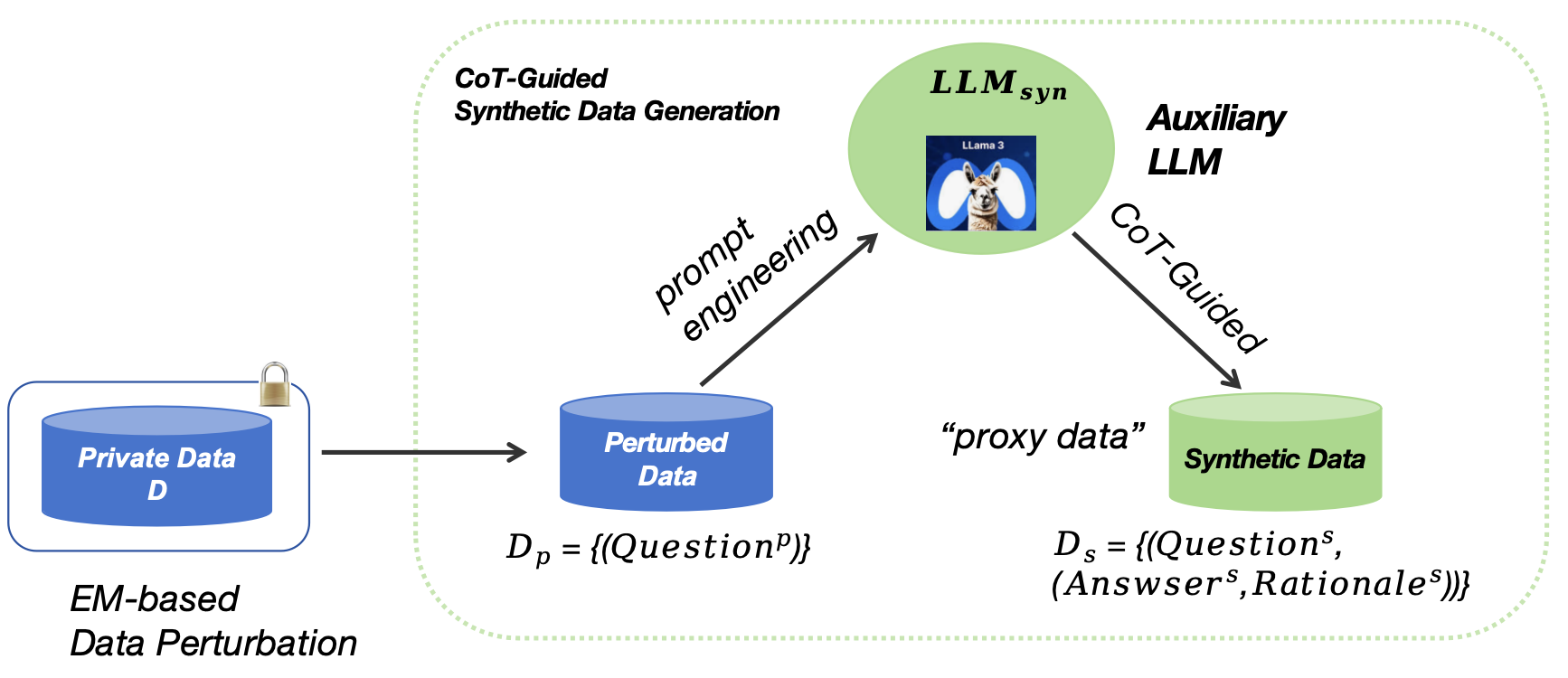}
  
    \caption{CoT-Guided 
Synthetic Data Generation.}
 \label{fig:syn_data}
\end{figure}

As illustrated in Figure \ref{fig:syn_data}, we introduce a simple and efficient method for generating synthetic data, utilizing prompt engineering techniques and CoT technology:

\begin{enumerate}
    \item \textbf{Question Generation.} We prompt $LLM_{syn}$ to create a new question, starting from a perturbed question. To enhance the validity of these new created questions, we enforce three guidelines within the prompt: (1) the new question needs to conform to common knowledge, (2) it must be solvable on its own, independent of the original question, and (3) it should not contain any answer responses. Furthermore, we establish specific formatting standards for both questions and answers, customized to suit the needs of various datasets~\cite{li2024common}.

    \item \textbf{Answer Generation.} We instruct $LLM_{syn}$ to generate a CoT response for every newly created question. For consistency, we request $LLM_{syn}$ to generate answers to the same question three times and check for agreement. If the answers differ, we reject the synthetic data.

    \item \textbf{Rationale Generation.} We request $LLM_{syn}$ to generate rationales for each synthetic data using the CoT prompting technique.
\end{enumerate}

Detailed prompt designs are presented in Appendix \ref{sec:appendix-prompt}. The generated synthetic data and their rationales are then employed for model pruning and retraining on the server-side.

\subsection{Rationale-Aware Structured Pruning} 
\label{sec:pruning}

LLMs exhibit layer-wise redundancy, with deeper layers often showing higher levels of functional overlap. To identify and remove redundant layers effectively, we need a quantitative metric that can assess each layer's contribution to the model's performance. This intrinsic metric should evaluate both the layer's individual importance and its interaction with other layers in maintaining the model's overall functionality.

To quantify the impact of each layer, we use a novel metric termed "Block Influence" (BI), which is proposed in the ShortGPT~\cite{men2024shortgpt}. This metric is grounded in the hypothesis that a transformer block's significance is directly proportional to the extent it modifies the hidden states. Mathematically, the BI score for the $i^{th}$ block is computed as:
\begin{align}
	\text{BI}_i = 1 - \mathbb{E}_{X,t} \left[ \frac{X_{i,t}^T X_{i+1,t}}{||X_{i,t}||_2 ||X_{i+1,t}||_2} \right],
\end{align}
where $X_i$ denotes the input to the $i^{th}$ layer, and $X_{i,t}$ represents the $t^{th}$ row of $X_i$.

On the server, we utilize the synthetic dataset $\mathcal{D}_s$, as described in Section \ref{sec:synthetic}, to compute the BI score for each layer of the $LLM_o$ model, denoted as $f_{\theta_o}$. This model represents the original LLM that requires pruning.

The original BI method~\cite{men2024shortgpt} relies solely on input and task label information, processed through a single forward pass: $f_{\theta}(x^s_i) \rightarrow y^s_i$.
\textit{\textbf{We further extend the BI computation to encompass two distinct facets of influence:}} $f_{\theta}(x^s_i) \rightarrow y^s_i$ and $f_{\theta}(x^s_i) \rightarrow r^s_i$. This enhancement not only facilitates the prediction of task labels but also enables the generation of corresponding rationales based on the inputs. 
\textbf{\textit{Our novel BI score is determined as follows:}}
\begin{align}
\label{eq:bi}
	\text{BI}_i = \text{BI}_{\text{Label},i} + \text{BI}_{\text{Rationale},i}
\end{align}
where $\text{BI}_{\text{Label},i}$ and $\text{BI}_{\text{Rationale},i}$ signify the influences pertaining to label predictions and rationale generation, respectively.

\begin{figure}[ht]
 \centering

  \includegraphics[width=1\linewidth]{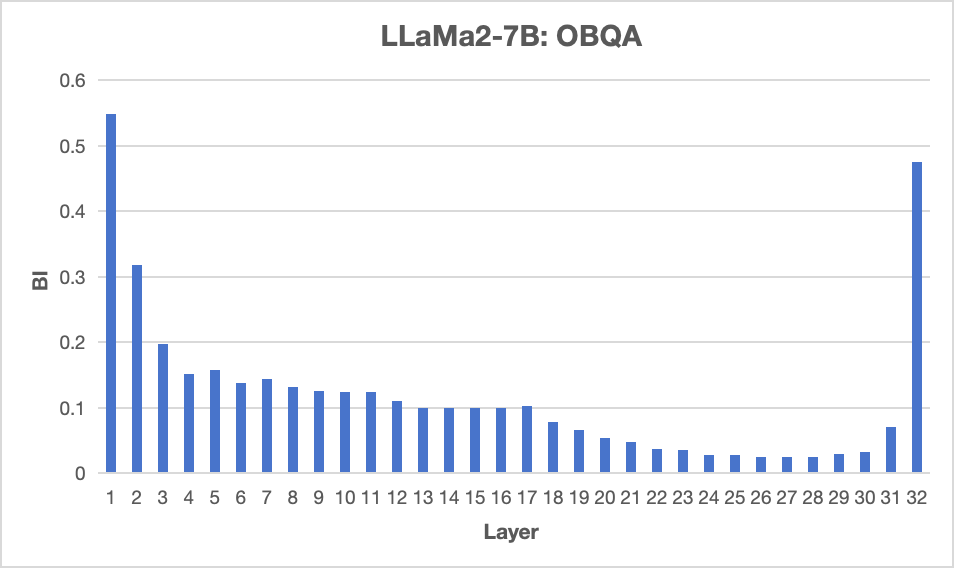}
  
    \caption{Layer Importance Example: The significance of each layer, as indicated by the BI (Block Influence) value of LLaMA2-7B on the OBQA dataset, based on the PPC-GPT framework.}
 \label{fig:bi_llama_obqa}
\end{figure}

A higher BI score indicates greater layer importance in the model architecture. As illustrated in Figure \ref{fig:bi_llama_obqa}, we leverage these BI scores to guide our pruning strategy: layers are arranged in ascending order based on their BI scores, and those with lower scores are systematically removed to obtain the pruned model structure SLM $f_{\phi}$.

\subsection{Retraining}
\label{sec:retraining}
We employ the term "retraining" to designate the process of performance recovery subsequent to pruning. 
In this section,  retraining is divided into two stages: (1) \textit{Server-side Retraining}, and (2) \textit{Client-side Retraining}.

\textbf{Server-side Retraining.}
On the server side, we utilize the synthetic dataset $\mathcal{D}_s$, as described in Section \ref{sec:synthetic}, to retrain the pruned model SLM $f_{\phi}$. \textbf{\textit{We propose CoT knowledge distillation, guided by rationales generated by $LLM_{syn}$, to enhance the performance of SLM $f_{\phi}$}}.
Formally, we conceptualize the learning process with rationales as a \textit{multi-task learning} problem~\cite{zhang2021survey, wei2022chain,hsieh2023distilling}. Specifically, we train the model $f_{\phi}(x^s_i) \rightarrow (y^s_i, r^s_i)$ to achieve not only the prediction of task labels but also the generation of corresponding rationales based on textual inputs. 
This multi-task training ensures that our model produces not only accurate predictions but also insightful justifications for its decisions, thereby enhancing the model's transparency and explainability. The multi-task learning objective can be formulated as follows:

\begin{equation}\label{eq:cot_kd}
\mathcal{L} = \mathcal{L}_{\text{Label}} + \mathcal{L}_{\text{Rationale}}
\end{equation}
where $\mathcal{L}_{\text{Label}}$ represents the label prediction loss:

\begin{equation}\label{eq:ft}
\mathcal{L}_{\text{Label}}(\phi; \mathcal{D}_s) = \mathbb{E}_{(x^s, y^s) \sim \mathcal{D}_s} \ell_{\text{CE}}(f_{\phi}(x^s), y^s)
\end{equation}
and $\mathcal{L}_{\text{Rationale}}$ represents the rationale generation loss:
\begin{equation}\label{eq:rationale}
\mathcal{L}_{\text{Rationale}}(\phi; \mathcal{D}_s) = \mathbb{E}_{(x^s, r^s) \sim \mathcal{D}_s} \ell_{\text{CE}}(f_{\phi}(x^s), r^s)
\end{equation}
where $\ell_{\text{CE}}$ denotes the cross-entropy loss.

\textbf{Client-side Retraining.}
On the client side, we utilize local private data $\mathcal{D}$ to further retrain the pruned model, SLM $f_{\phi}$, once it has been received from the server. \textit{\textbf{Our work encompasses conventional training, leveraging ground truth labels to further enhance the performance of SLM $f_{\phi}$}}.
Formally,  the label prediction loss for this dataset $\mathcal{D}$ is formulated as follows:
\begin{equation}\label{eq:ft_client}
\mathcal{L}_{\text{Label}}(\phi; \mathcal{D}) = \mathbb{E}_{(x, y) \sim \mathcal{D}} \ell_{\text{CE}}(f_{\phi}(x), y)
\end{equation}

\section{Experiments}

\subsection{Setup}
We have devised a scenario to assess the performance of the PPC-GPT framework across various text generation tasks. This setup employs a client-server architecture, where the server hosts an \textit{auxiliary LLM} for synthetic data generation, denoted as $LLM_{syn}$. Specifically, we have selected LLaMA3-70B~\cite{dubey2024llama} for this purpose. For model pruning, we utilize LLaMA2-7B~\cite{touvron2023llama2} and OPT-6.7B~\cite{zhang2022opt} as the source models, denoted as $LLM_o$.  In the default setting, the privacy budget $\epsilon=3$,  and the synthetic data ratio is 8.

\textbf{Datasets and Evaluation Metrics}. We conduct a comparative evaluation of PPC-GPT on QA datasets. Specifically, we include CommonsenseQA (CQA)~\cite{talmor2019commonsenseqa}, OpenBookQA (OBQA)~\cite{mihaylov2018can}, ARC-C~\cite{clark2018think}, ARC-E~\cite{clark2018think}, FiQA-SA~\cite{maia201818}. For these datasets, we primarily use \textbf{Accuracy} as the evaluation metric. It's worth noting that in our experiments, all methods undergo zero-shot evaluation and we use the \textit{lm-evaluation-harness} package~\cite{eval-harness}.

\textbf{Baselines}. 
To evaluate the performance of our PPC-GPT framework, we conducted a comparative analysis against the following baselines:

\begin{itemize}
    \item DenseSFT, where the client independently  fine-tunes $LLM_o$ using its private dataset.

     \item Plain-C, where the client independently prunes $LLM_o$ using its private dataset (suppose the client can deploy $LLM_o$) and subsequently fine-tunes the pruned model.

    \item DP-Instruct-C~\cite{yu2024privacy}, where the client finetunes generator (e.g., LLaMA2-1.3B) with DP-SGD and using synthetic datasets generated from generator to prune $LLM_o$ and subsequently fine-tunes the pruned model with the private dataset.

\end{itemize}

\subsection{Main Results}

In our experiments, we extensively evaluated the performance of the proposed PPC-GPT framework across various text generation tasks. Notably, given that current structured pruning methods typically reduce parameters by no more than 30\%, we conducted experiments with approximately 30\% of the parameters pruned. Additional experiments exploring different parameter reduction proportions will be discussed in Section \ref{sec:exp-prune-ratio}.

As shown in Table \ref{tab:compare_method}, the results highlight  the effectiveness of PPC-GPT in compressing LLMs into task-specific SLMs while prioritizing data privacy protection, when compared to other baseline approaches.   PPC-GPT outperforms the DP-Instruct-C method, which utilizes DP-SGD for privacy protection during model compression. Furthermore, PPC-GPT even surpasses the Plain-C method, which directly compresses the model using private data.  Additionally, when compared to DenseSFT, the compressed model in PPC-GPT even outperforms the raw model on some datasets.
Specifically,  taking LLaMA2-7B for an example, in the LLaMA2-7B model, 
PPC-GPT outperforms the DP-Instruct-C method by 0.4\%,  5.2\%, 5\%,  15.1\%, and 1.8\% on the CQA, OBQA, ARC-E, ARC-C, and FiQA-SA datasets, respectively.  
Similarly, PPC-GPT exceeds the Plain-C method by 0.7\%,  2\%, 4.5\%,  8.2\%, and 1.6\% on the respective datasets.

\begin{table*}[ht]
    \centering
    \footnotesize
    \begin{tabular}{lccccccc}
        \toprule
        & & &  \multicolumn{5}{c}{\textbf{DataSets}} \\
        \cmidrule(lr){4-8}
        \textbf{Model} & \textbf{Method} & \textbf{Ratio (\%)} & \textbf{CQA} & \textbf{OBQA} & \textbf{ARC-E} & \textbf{ARC-C} & \textbf{FiQA-SA} \\
        \midrule
        \multirow{5}{*}{LLaMA2-7B} & DenseSFT& 0 & $81.6_{\pm 0.54}$ & $80.3_{\pm 0.50}$ & $82.9_{\pm 0.18}$ & $60.0_{\pm 0.42}$ & $68.9_{\pm 1.66}$ \\

        \cmidrule(lr){2-8}
        & Plain-C& 30 &  $77.6_{\pm 0.14}$  &  $77.9_{\pm 0.16}$ & $79.7_{\pm 0.29}$ & $54.0_{\pm 0.82}$  & $71.1_{\pm 1.37}$\\
        
        \cmidrule(lr){2-8}
        & DP-Instruct-C&30 & $77.9_{\pm 0.62}$ & $74.7_{\pm 1.32}$ & $79.2_{\pm 0.33}$ & $47.1_{\pm 4.10}$ & $70.9_{\pm 0.83}$ \\
        
        \cmidrule(lr){2-8}
        & \textbf{PPC-GPT} & 30 & $78.3_{\pm 0.41}$  & $79.9_{\pm 0.57}$ & 
        $84.2_{\pm 0.33}$  & $62.2_{\pm 0.61}$ & $72.7_{\pm 0.54}$\\

        \midrule
        \multirow{5}{*}{OPT-6.7B}& DenseSFT& 0 & $75.4_{\pm 0.64}$ & $60.0_{\pm 0.99}$ & $65.8_{\pm 0.70}$ & $31.4_{\pm 0.86}$ & $70.0_{\pm 1.09}$ \\

         \cmidrule(lr){2-8}
        & Plain-C& 30 & $47.4_{\pm 1.12}$ & $36.5_{\pm 1.48}$  &  $40.2_{\pm 0.89}$ & $27.6_{\pm 0.37}$  & $52.4_{\pm 1.37}$\\
        
        \cmidrule(lr){2-8}
        & DP-Instruct-C&30 &  $58.7_{\pm 2.04}$ & $39.7_{\pm 1.04}$ & $44.5_{\pm 2.53}$  & $28.6_{\pm 1.72}$  & $54.5_{\pm 1.67}$\\
       
        \cmidrule(lr){2-8}
        & \textbf{PPC-GPT} & 30 & $65.6_{\pm 0.95}$ & $52.1_{\pm 0.96}$ & $57.3_{\pm 0.16}$ & $36.0_{\pm 0.59}$ & $64.9_{\pm 1.26}$\\
    
        \bottomrule
    \end{tabular}
    \caption{Performance Comparison of Compression Methods on LLMs.}
    \label{tab:compare_method}
\end{table*}

\subsection{Ablation Study}

\subsubsection{Impact of Different Privacy Budgets}

In this section, we explore the impact of privacy budgets on the performance of PPC-GPT. Table \ref{tab:compare_privacy} presents PPC-GPT's performance across a range of privacy budgets ($\epsilon = 1, 3, 5, 10$). 
Notably, when juxtaposed with Table \ref{tab:compare_method}, it becomes apparent that even with a privacy budget of $\epsilon=1$, PPC-GPT outperforms the Plain-C method by 1.7\% and 3.4\% on the OBQA and ARC-E datasets, respectively, within the LLaMA2-7B model. Similarly, PPC-GPT exceeds it by 14\% and 14.4\% in the OPT-6.7B model. 
As the privacy budget $\epsilon$ increases, PPC-GPT's performance demonstrates a significant improvement, highlighting its proficiency and adaptability in achieving a balance between privacy and utility.

\begin{table}[ht]  
    \centering  
    \footnotesize
    \setlength{\tabcolsep}{3.5 pt}
    \begin{tabular}{lcccccc}  
        \toprule  
        & & & \multicolumn{4}{c}{\textbf{Privacy Budget($\epsilon$)}} \\  
        \cmidrule(lr){4-7}  
        \textbf{Model} & \textbf{Datasets} & \textbf{Stage}& \textbf{1} & \textbf{3} & \textbf{5} & \textbf{10} \\  
        \midrule  
        \multirow{4}{*}{LLaMA2}& \multirow{2}{*}{OBQA}   
                & S& 65.4 & 67.1 & 67.9 & 69.4 \\  
            \cmidrule(lr){3-7}  
            & & C& 79.6 & 79.9 & 80.1 & 79.8 \\  
            \cmidrule(lr){2-7}  
            & \multirow{2}{*}{ARC-E}   
                & S& 78.8 & 80.4 & 79.9 & 79.5 \\  
            \cmidrule(lr){3-7}  
            & & C& 83.1 & 84.2 & 84.4 & 83.4 \\  
        \midrule  
        \multirow{4}{*}{OPT}& \multirow{2}{*}{OBQA}   
                & S& 35.7 & 36.3 & 36.1 & 38.8 \\  
            \cmidrule(lr){3-7}  
            & & C& 50.5 & 52.1 & 52.4 & 53.5 \\  
            \cmidrule(lr){2-7}  
            & \multirow{2}{*}{ARC-E}   
                & S& 49.1 & 50.4 & 49.3 & 50.5 \\  
            \cmidrule(lr){3-7}  
            & & C& 54.6 & 57.3 & 55.5 & 55.3 \\  
        \bottomrule  
    \end{tabular}  
    \caption{Comparison of PPC-GPT's performance across \textbf{different privacy budgets $\epsilon$}.  \textbf{S} denotes the performance of  target SLM on the server-side, while \textbf{C} represents the performance of target SLM on the client-side.}
    \label{tab:compare_privacy}
\end{table}

\subsubsection{Impact of Different Synthetic Data}

In this section, we explore the impact of synthetic data on PPC-GPT's performance, considering two dimensions: the synthetic data ratio and the inclusion of rationales in synthetic data.

\textbf{Synthetic Data Ratio}. 
Table \ref{tab:compare_synthetic_data_ratio} presents the performance of PPC-GPT across various synthetic data ratios (ratio = 1, 2, 4, 8). 
As the ratio of synthetic data increases, PPC-GPT's performance exhibits a substantial improvement, highlighting the crucial role of the synthetic data ratio and indicating that a higher amount of synthetic data results in further improvements.
Specifically, PPC-GPT with the synthetic data ratio of 8 outperforms the ratio of 1 by 1.7\% and 4.1\% on the OBQA and ARC-E datasets, respectively, within the LLaMA2-7B model. Similarly, with the OPT-6.7B model, it exceeds the ratio of 1 by 4.2\% and 7.6\%.

\textbf{Synthetic Data Rationales}.
We undertake an analysis to investigate the effects of rationales on PPC-GPT's performance. Table \ref{tab:compare_rationales} compares PPC-GPT's performance between synthetic data with and without rationales (PPC-GPT w/ rationales and PPC-GPT w/o rationales). 
The findings demonstrate that PPC-GPT exhibits superior performance when the rationales of synthetic data is utilized, as compared to when it is absent.
Specifically, PPC-GPT w/ rationales outperforms PPC-GPT w/o rationales by 0.8\% and 0.9\% on the OBQA and ARC-E datasets, respectively, within the LLaMA2-7B model. Similarly, with the OPT-6.7B model, PPC-GPT w/ rationales exceeds PPC-GPT w/o rationales by 7\% and 9.1\%.

\begin{table}[ht]  
    \centering  
    \footnotesize
    \setlength{\tabcolsep}{3.5pt}
    \begin{tabular}{lcccccc}  
        \toprule  
        & & & \multicolumn{4}{c}{\textbf{Synthetic Data Ratio}} \\  
        \cmidrule(lr){4-7}  
        \textbf{Model} & \textbf{Datasets} & \textbf{Stage}& \textbf{1} & \textbf{2} & \textbf{4} & \textbf{8} \\  
        \midrule  
        \multirow{4}{*}{LLaMA2}   
            & \multirow{2}{*}{OBQA}   
                & S& 62.3 & 64.6 & 64.6 & 67.1 \\  
            \cmidrule(lr){3-7}  
            & & C& 78.2 & 78.3 & 78.5 & 79.9 \\  
            \cmidrule(lr){2-7}  
            & \multirow{2}{*}{ARC-E}   
                & S& 73.5 & 75.5 & 77.9 & 80.4 \\  
            \cmidrule(lr){3-7}  
            & & C& 80.1 & 80.8 & 82.3 & 84.2 \\  
        \midrule  
        \multirow{4}{*}{OPT}   
            & \multirow{2}{*}{OBQA}   
                & S& 32.9 & 34.7 & 36.9 & 36.3 \\  
            \cmidrule(lr){3-7}  
            & & C& 47.9 & 50.2 & 51.5 & 52.1 \\  
            \cmidrule(lr){2-7}  
            & \multirow{2}{*}{ARC-E}   
                & S& 40.4 & 43.9 & 47.5 & 50.4 \\  
            \cmidrule(lr){3-7}  
            & & C& 49.7 & 52.3 & 54.9 & 57.3 \\  
        \bottomrule  
    \end{tabular}  
    \caption{
    Comparison of PPC-GPT's performance across \textbf{different synthetic data ratio}.}
    \label{tab:compare_synthetic_data_ratio}
\end{table}

\begin{table}[ht]  
    \centering  
    \footnotesize
    \begin{tabular}{lcccc}  
        \toprule  
        & & & \multicolumn{2}{c}{\textbf{Rationales}} \\  
        \cmidrule(lr){4-5}  
        \textbf{Model} & \textbf{Datasets} & \textbf{Stage}& \textbf{w/} & \textbf{w/o} \\  
        \midrule  
        \multirow{4}{*}{LLaMA2}   
            & \multirow{2}{*}{OBQA}   
                & S& 67.1 & 65.9 \\  
            \cmidrule(lr){3-5}  
            & & C& 79.9 & 79.1 \\  
            \cmidrule(lr){2-5}  
            & \multirow{2}{*}{ARC-E}   
                & S& 80.4 & 77.9 \\  
            \cmidrule(lr){3-5}  
            & & C& 84.2 & 83.3 \\  
        \midrule  
        \multirow{4}{*}{OPT}& \multirow{2}{*}{OBQA}   
                & S& 36.3 & 31.1 \\  
            \cmidrule(lr){3-5}  
            & & C& 52.1 & 45.1 \\  
            \cmidrule(lr){2-5}  
            & \multirow{2}{*}{ARC-E}   
                & S& 50.4 & 43.2 \\  
            \cmidrule(lr){3-5}  
            & & C& 57.3 & 48.2 \\  
        \bottomrule  
    \end{tabular}  
    \caption{
    Comparison of PPC-GPT's performance: with vs. without \textbf{rationales}.}  
    \label{tab:compare_rationales}  
\end{table}

\subsubsection{Impact of Server-Side Retraining}
In this section, we explore the impact of server-side retraining on the performance of PPC-GPT. Table \ref{tab:compare_retrain} presents a comparison of PPC-GPT's performance with and without server-side retraining.
The findings demonstrate that PPC-GPT exhibits superior performance when server-side retraining is utilized, as compared to when it is absent.
Specifically, PPC-GPT w/ server-side retraining outperforms PPC-GPT w/o server-side retraining by 2\% and 4.5\% on the OBQA and ARC-E datasets, respectively, within the LLaMA2-7B model. Similarly, with the OPT-6.7B model, PPC-GPT w/ server-side retraining exceeds PPC-GPT w/o server-side retraining by 15.1\% and 15.7\%.

\begin{table}[ht]
    \centering
    \footnotesize
    \begin{tabular}{llcc}
        \toprule
        & & \multicolumn{2}{c}{\textbf{Server:Retraining}} \\
        \cmidrule(lr){3-4}
        \textbf{Model} & \textbf{Dataset} & \textbf{w/} & \textbf{w/o}  \\
        \midrule
        \multirow{2}{*}{LLaMA2} & OBQA & $79.9$ & $77.9$ \\
        \cmidrule(lr){2-4}
        & ARC-E & $84.2$ & $79.7$  \\
        \midrule
        \multirow{2}{*}{OPT}& OBQA & $52.1$ & $37.0$  \\
         \cmidrule(lr){2-4}
        & ARC-E & $57.3$ & $41.6$ \\
        \bottomrule
    \end{tabular}
    \caption{Comparison of PPC-GPT's performance: with vs. without \textbf{server-side retraining}.}  
    \label{tab:compare_retrain}  
\end{table}

\subsubsection{Impact of Different Importance Metric} 

In this section, we explore the impact of different important metrics on PPC-GPT’s performance:

\textbf{Seq}: The importance is directly correlated with the sequence order, where the shallower layers hold greater importance. 

\textbf{BI}: BI mentioned in previous section \ref{sec:pruning}.

Table \ref{tab:compare_importance} presents PPC-GPT's performance across different important metrics. 
The findings demonstrate that PPC-GPT with BI exhibits superior performance than PPC-GPT with Seq. 

\begin{table}[ht]  
    \centering  
    \footnotesize
    \begin{tabular}{lcccc}  
        \toprule  
        & & & \multicolumn{2}{c}{\textbf{Important}} \\  
        \cmidrule(lr){4-5}  
        \textbf{Model} & \textbf{Datasets} & \textbf{Stage}& \textbf{BI}& \textbf{Seq}\\  
        \midrule  
        \multirow{4}{*}{LLaMA2}   
            & \multirow{2}{*}{OBQA}   
                & S& $67.1$ & $66.5$\\  
            \cmidrule(lr){3-5}  
            & & C& $79.9$ & $79.9$\\  
            \cmidrule(lr){2-5}  
            & \multirow{2}{*}{ARC-E}   
                & S& $80.4$ & $80.0$ \\  
            \cmidrule(lr){3-5}  
            & & C&  $84.2$ &  $83.9$\\  
        \midrule  
        \multirow{4}{*}{OPT}& \multirow{2}{*}{OBQA}   
                & S& $36.3$ & $34.7$\\  
            \cmidrule(lr){3-5}  
            & & C& $52.1$ & $48.3$ \\  
            \cmidrule(lr){2-5}  
            & \multirow{2}{*}{ARC-E}   
                & S& $50.4$ & $43.7$ \\  
            \cmidrule(lr){3-5}  
            & & C& $57.3$ &  $51.7$ \\  
        \bottomrule  
    \end{tabular}  
    \caption{Comparison of PPC-GPT's performance across \textbf{different importance metrics}.}  
    \label{tab:compare_importance}  
\end{table}

\subsubsection{Impact of Different Model Pruning Ratio} 
\label{sec:exp-prune-ratio}
In this section, we explore the impact of different model pruning ratio on PPC-GPT's performance.
Table \ref{tab:compare_pruner_ratio} presents the performance of PPC-GPT across different model pruning ratios (namely, 0\%, 30\%, 50\%, and 70\%). As the pruning ratio increases, the performance of PPC-GPT exhibits a decline.

\begin{table}[ht]
    \centering
    \footnotesize
    \begin{tabular}{lccc}
        \toprule
        & &  \multicolumn{2}{c}{\textbf{DataSets}} \\
        \cmidrule(lr){3-4}
        \textbf{Model}  & \textbf{Ratio (\%)} & \textbf{OBQA} & \textbf{ARC-E} \\
        \midrule
        \multirow{4}{*}{LLaMA2} & 0  & $80.3$ & $82.9$ \\
        
        \cmidrule(lr){2-4}
        & 30  & $79.9$ & $84.2$   \\

        \cmidrule(lr){2-4}
        & 50 & $74.4$ &  $76.8$\\

        \cmidrule(lr){2-4}
        & 70 & $35.3$ & $37.4$ 
         \\
        
        \midrule
        \multirow{4}{*}{OPT}&  0  & $60.0$ & $65.8$ \\
        
        \cmidrule(lr){2-4}
         & 30  & $52.1$ & $57.3$  \\
       
        \cmidrule(lr){2-4}
        & 50 & $36.1$  & $38.3$  \\
        
        \cmidrule(lr){2-4}
        & 70 & $30.9$ & $33.2$ \\
        
        \bottomrule
    \end{tabular}
    \caption{Comparison of PPC-GPT's performance across \textbf{different pruning ratios}.}
    \label{tab:compare_pruner_ratio}
\end{table}

\section{Conclusions}

In this paper, we introduced PPC-GPT, a novel federated framework for compressing LLMs into task-specific SLMs while preserving privacy. Our framework integrates four key components: exponential mechanism-based data perturbation, CoT-guided synthetic data generation, rationale-aware structured pruning, and CoT-based knowledge distillation. Experiments demonstrate that PPC-GPT effectively compresses LLMs while maintaining comparable performance and ensuring privacy protection. This work provides a practical solution for deploying LLMs in resource-constrained, privacy-sensitive scenarios.

\section*{Limitations}
While PPC-GPT shows promising results in compressing LLMs into task-specific SLMs while ensuring data privacy, it has several limitations. Firstly, PPC-GPT relies on an auxiliary LLM with robust CoT capabilities to generate high-quality synthetic data and rationales. These synthetic data are crucial for guiding the structured pruning and retraining processes of both the source LLM (the model slated for compression) and the target SLM (the compressed model). If the auxiliary LLM lacks sophisticated CoT reasoning abilities, the quality and diversity of the generated synthetic data may be compromised, which in turn could adversely affect the performance of the compressed SLMs. This limitation underscores the importance of selecting or pre-training an auxiliary LLM with strong CoT capabilities when deploying PPC-GPT. However, it's important to note that the source LLM (the model slated for compression) does not necessarily require CoT capabilities. Furthermore, as observed in our experiments, the performance of PPC-GPT tends to degrade with higher pruning ratios. This indicates that optimizing the pruning strategy to strike a better balance between model size and performance remains an open challenge.

\bibliography{ref}

\appendix

\section{Privacy Analysis of PPC-GPT}
\label{sec:appendix-privacy}
 Our privacy protection strategy in PPC-GPT is grounded in rigorous theoretical foundations and validated through comprehensive empirical studies. The framework implements a theoretically-sound differential privacy (DP) mechanism that operates at the token-level feature space, completely eliminating the need for raw data transmission. Specifically, we adopt the exponential mechanism, which provides formal $\epsilon$-DP guarantees and has been extensively analyzed in privacy-preserving NLP literature \cite{yue2021differential, chen-etal-2023-customized, tong2025inferdpt}. The theoretical privacy guarantees of this mechanism are well-established, allowing us to focus on its practical implementation and performance optimization rather than re-establishing its privacy properties.

\section{Implementation Details}
\label{sec:appendix-impl}
\subsection{Hyperparameter Settings}

During the training process, we specifically configured the parameters. Specifically, we set the batch size to 32 and utilized the AdamW optimizer. The maximum number of training steps varied between 300 and 6400. Additionally, we established a learning rate of 5e-5. For the input and target lengths, we set the maximum question length to 64 and the maximum target length to 128.
For the LoRA configuration of LLaMA2, we set the LoRA alpha to 32 and the LoRA rank to 8. In contrast, for the OPT model, we configured the LoRA alpha to 64 and the LoRA rank to 32. The Lora dropout for both models was set to 0.1.

\subsection{Data Splitting}

For the datasets, all splits (training, validation, and test) were downloaded from HuggingFace \cite{lhoest-etal-2021-datasets}.

\subsection{Dataset Licenses}
All the datasets were downloaded from HuggingFace\cite{lhoest-etal-2021-datasets} and under Apache License, Version 2.0.

\subsection{Machine Configuration}

The experiments were conducted on machines equipped with 4 and 8 Nvidia V100 32G.

\section{Synthetic Prompt Templates}
\label{sec:appendix-prompt}
Table \ref{tab:prompt1} and \ref{tab:prompt2} provide prompt templates for question generation, answer generation, and rationale generation.

\begin{table*}[htbp]
    \centering
    \begin{tabular}{p{2cm} | p{14cm}}
    \toprule
    {Tasks} & {Prompts} \\
    \midrule
    Question \indent Generation & \begin{minipage}[t]{14cm}
    \vspace{0pt} 
Please act as a professional teacher.

Your goal is to promote research in advanced question-answering, probing a deeper understanding of both the topic (with salient facts summarized as an open book, also provided with the dataset) and the language it is expressed in.

You will be given a multiple-choice question. Please create a new question and multiple choices based on the Given Question And Multiple Choices and following instructions.

To achieve the goal, you have two jobs.

\# Please generate a similar but new question and multiple choices according to the Given Question And Multiple Choices.

\# Check the question and multiple choices by solving it step-by-step to find out if it adheres to all principles.

You have eight principles to do this.

\# Ensure the new question only asks for one thing, be reasonable, be based on the Given Question And Multiple Choices, and can be answered with only one right choice.

\# Ensure the new questions requires multi-step reasoning, use of additional common and commonsense knowledge.

\# Ensure the new question is in line with common sense of life.

\# Ensure your student can answer the new question without the given question. If you want to use some numbers, conditions or background in the given question, please restate them to ensure no information is
omitted in your new question.

\# Please DO NOT include solution in your question.

\# Make sure the choices in CREATED QUESTION AND CHOICES are list format, starts with [ and ends with ].

\# Ensure only one choice in CREATED QUESTION AND CHOICES is right.

\# Ensure your output only has three lines, the first line is "CREATED QUESTION AND CHOICES:", the second line starts with "Question:", and the third line starts with "Choices".

Given Question and Multiple Choices:
\{question\}, 
\{choices\}

Your output should be in the following format:

CREATED QUESTION AND CHOICES:

Question: <your created question>

Choices: <your created choices>
\\
\end{minipage}
\\
    \hline
    Answer \indent Generation & \begin{minipage}[t]{14cm}
    \vspace{0pt} 

    Please act as a professional teacher.
    
    Your goal is to accurately solve a multiple-choice question.
    
    To achieve the goal, you have two jobs.
    
    \# Write detailed solution to a Given Question.
    
    \# Write the final choice to this question.
    
    You have three principles to do this.
    
    \# Ensure the solution is step-by-step.
    
    \# Ensure the final answer is just a letter. 
    
    \# Use of additional common and commonsense knowledge.
    
    Given Question and Choices:
    \{question\},
    \{choices\}
    
    Your output should be in the following format: 
    
    SOLUTION: <your detailed solution to the given question>
    
    FINAL ANSWER: <your final choice to the question with only an uppercase letter>
\\
\end{minipage}
    \\

\bottomrule
         
    \end{tabular}
    \caption{The prompt templates are used for generating questions and answers.
    }
   
    \label{tab:prompt1}
\end{table*}

\begin{table*}[htbp]
    \centering
    \begin{tabular}{p{2cm} | p{14cm}}
    \toprule
    {Tasks} & {Prompts} \\
    \midrule
    
    Rationale \indent Generation & \begin{minipage}[t]{14cm}
    \vspace{0pt} 

    You are given the right Answer from Choices, please explain it in "Rationale" with few words. Please refer to the example to write the rationale. 
    
    Try to generate logically clear and correct rationale. Reply in english only and use '<end>' to finish your rationale. Your reply format must strictly follow the provided example and reply rationale contents only!
    
    Example(s):
    
    Question: The sun is responsible for
    
    Choices: ['puppies learning new tricks', 'children growing up and getting old', 'flowers wilting in a vase', 'plants sprouting, blooming and wilting']
    
    Answer: 'plants sprouting, blooming and wilting'.
    
    Rationale: The sun provides light and warmth, essential for the process of photosynthesis in plants, which enables them to grow, bloom, and eventually wilt due to natural life cycles.
    
    Please explain:
    
    Question: \{question\}
    
    Choices: \{choices.text\}
    
    Answer: \{choices.text[choices.label.index(answerKey)]\}
    
    Rationale:
    \\
   \end{minipage}
 \\
    
\bottomrule
         
    \end{tabular}
    \caption{The prompt templates are used for generating rationale.
    }
   
    \label{tab:prompt2}
\end{table*}

\end{document}